\pdfoutput=1 
\documentclass[11pt]{article}
\usepackage{latexsym} 
\usepackage{tikz} 
\usepackage{pgfplots}
\usepackage[utf8]{inputenc}
\usepackage[T1]{fontenc}
\usepackage{graphicx}
\usepackage{comment}
\usepackage{enumitem}
\usepackage[normalem]{ulem}
\usepackage{amsmath}
\usepackage{textcomp}
\usepackage{amssymb}
\usepackage{capt-of}
\usepackage{hyperref}
\usepackage{EMNLP2023}
\usepackage{times}
\usepackage{todonotes}
\usepackage{covington}
\usepackage{color,soul}
\sethlcolor{white} 
\newcommand{\td}[1]{\todo[color=yellow]{\scriptsize #1\par}}

\title{BERT Goes Off-Topic: Investigating the Domain Transfer Challenge using Genre Classification}

\author{Dmitri Roussinov\\
  University of Strathclyde\\16 Richmond Street, Glasgow G1 1XQ\\
  \texttt{dmitri.roussinov@strath.ac.uk} \\\And
  Serge Sharoff \\
  University of Leeds\\ Leeds LS2 9JT\\
  \texttt{s.sharoff@leeds.ac.uk}\\}

\begin{document}

\maketitle
\begin{abstract}
While performance of many text classification tasks has been recently improved due to Pre-trained Language Models (PLMs), in this paper we show that they still suffer from a performance gap when the underlying distribution of topics changes.  For example, a genre classifier trained on \textit{political} topics often fails when tested on documents about \textit{sport} or \textit{medicine}. In this work, we quantify this phenomenon empirically with a large corpus and a large set of topics. Consequently, we verify that domain transfer remains challenging both for classic PLMs, such as BERT, and for modern large models, such as GPT-3. 
We also suggest and successfully test a possible remedy:
after augmenting the training dataset with topically-controlled synthetic texts, the F1 score improves by up to 50\% for some topics, nearing on-topic training results, while others show little to no improvement. While our empirical results focus on genre classification, our methodology is applicable to other classification tasks such as gender, authorship, or sentiment classification.
\end{abstract}

\section{Introduction}
\label{sec:orga362c94}

Automatic text classification is a critical task in natural language processing, enabling proper understanding, summarization, archiving, and retrieval of documents across various domains, such as legal and medical. This task has been greatly improved due to pre-trained language models
such as BERT \cite{devlin2018bert}, T5 \cite{raffel20} or GPTs \cite{brown2020language}. 
To achieve true artificial general intelligence (AGI), it is essential that trained computer models can recognize various document categories across different domains. However it has been noticed \cite{hendrycks20pretrained,moon2021masker} that while in general PLMs are more robust than previous models, they still suffer from spurious domain-specific clues. 
While all our methods proposed here apply to other non-topical text classification tasks such as sentiment or authorship identification, in this particular work we have taken a thorough look at \textit{document genre classifiers}: distinguishing between different styles (genres) of texts, such as academic articles, experimental protocols, regulatory documents, and patient leaflets \cite{santini10genreintro,sharoff10lrec}.  People can easily recognize document genres from just a few examples even if those examples are from a different domain \cite{crowston10}.  

Text classification research often contrasts the properties of topic vs. those of style  \cite{dewdney01}.  However, this contrast is difficult to maintain, as the training sets in most corpora for style or genre prediction are biased with respect to topics specific to individual styles or genres, so that classifiers do not transfer across corpora in case of variation between their topics. For example, a model {identifying} FAQs can learn to pay attention to such keywords as \emph{hurricane} and \emph{tax advice} in case these topics are common for FAQs in a specific training corpus \cite{sharoff10lrec}. 

So far, this cross-influence of topics and styles has not been studied in the context of PLMs such as BERT \cite{devlin2018bert}, T5 \cite{raffel20} or GPTs \cite{brown2020language}. 
There has also been no quantification of the gap in transferring genre/style classifiers to new domains. For instance, no study has yet assessed the performance degradation when a classifier is trained on \textit{political} topics but tested on texts about \textit{sports} or \textit{medicine}.

In light of the aforementioned challenges, our study offers the following novel contributions\footnote{The tools and the experimental setups are available at \url{https://github.com/dminus1/genre}}:
\begin{itemize}[noitemsep]
\item While our study primarily focuses on genre classification, the \textbf{methodology} we use to assess and mitigate domain transfer gaps can be broadly applied, making it suitable for other non-topical classifications such as authorship or sentiment identification;  
\item We have created a large corpus with ``\textbf{natural genre annotation}'' covering a range of topics with some biases;
\item We empirically \textbf{quantify} the domain transfer gap on our corpus, demonstrating drops in F1 classification performance by 20-30 absolute percentage points; 
\item We propose a data \textbf{augmentation} approach which involves training text generators that can produce synthetic documents in any of the \textbf{genres} present in the genre training corpus and on any \textbf{topic}, out of those identified by neural topic-modeling algorithm \cite{dieng20topic} trained on an unrelated topically diverse large corpus. 
\item We verify that augmenting the training dataset with synthetics texts generated by our approach \textbf{facilitates} domain transfer by improving F1 classification metric by 2-6 absolute percentage points in average and on some topics as much as from 57.6 to 73.0. 
This improvement surpasses a general data augmentation baseline that generates synthetic documents but does not apply any domain transfer mechanisms that we propose here.
\item Through \textbf{ablation studies}, we verify that all the components of our augmentation approach are crucial. Also, by varying hyper-parameters, we can identify the optimal augmentation setup and avoid performance degradation.
\item \hl{Through a qualitative exploratory study with \textbf{ChatGPT} we were able to confirm that even a much larger language model can still suffer from a domain transfer gap.}
\end{itemize}

\section {Related studies and baselines} 

There have been studies that looked at impact of out-of-domain training data on PLM-based classifiers.  In particular, \newcite{hendrycks20pretrained} noticed that while in general PLMs are more robust than previous models, they still suffer from spurious clues. However, they tested the transfer gap only on a few hand-picked datasets with similar tasks but different data distributions (e.g. sentiment analysis trained on \textit{book} reviews  applied to \textit{movie} reviews), while here we are presenting an original methodology based on a neural topic model to investigate domain transfer between a wide variety of topics. Also, none of the prior works looked at domain transfer for genre/style classification tasks which we do here.

Within the broader context of domain transfer, genre classification holds a unique position.  Automatic genre classification has been recognised as an important task since the 1990s \cite{roussinov2001genre,santini10genreintro}.  The effect of topical biases has been estimated empirically by considering the reduction in performance of genre classifiers across topics in the New York Times corpus \cite{petrenz10}.  

Several studies have also demonstrated the success of PLMs with respect to the genre classification tasks \cite{ronnqvist21multilingual,kuzman22ginco}.  However, there have been no studies of topical biases for these models.  The split between topics and styles has been studied for a related task, including disentangled representation \cite{john19disentangled} and other methods of topic-style decomposition \cite{romanov19,subramanian19style}.  However, our study focuses on the numerical estimates of the topic transfer gap on large samples diverse in topics and in genres.  

A related research area concerns the use of causal models for interpreting the biases of neural predictions, for example, with respect to gender \cite{vig20causal}.  There have been studies to investigate biases in neural models by adding counter-factuals \cite{hallmaudslay19gender,kaushik20causal}. 

It has been noted that well-established data augmentation (DA) methods in domains such as computer vision and speech recognition \cite{anaby2020not,giridhara2019study,krizhevsky2017imagenet}, relying on simple transformations of existing samples, cannot be easily applied to natural text since they can lead to syntactic and semantic distortions. For a survey of  DA approaches for various natural language processing tasks we refer a reader to \newcite{feng2021survey}. 
The survey mentions several studies showing that DA is generally much less beneficial when applied to out-of-domain data (as studied here), likely because ``the distribution of augmented data can substantially differ from the original data." While  only a few of the surveyed works involved PLMs, the survey points out that PLMs have made many previously useful DA techniques \textbf{obsolete} since fine-tuned PLM-based classifiers already achieve high performance, as they have been pre-trained on large and diverse corpora. 
For those reasons, we decided not to contrast our approach with any of the classical pre-PLM domain transfer techniques, such as \newcite{blitzer07} or \newcite{daume10frustratingly}.  

While up to our knowledge, none of the prior works has specifically looked into the domain transfer gap for genre (or style) classification, 
it is still worth to note several closely related works, some of them included in \newcite{feng2021survey} {survey that involved PLMs not only as classifiers but also as generators for augmented data. 
This includes \newcite{kumar2020data} who looked at sentiment/intent/question  classification, \newcite{lee2021neural} who targeted under-represented categories, \newcite{edwards2021guiding} who focused on selecting the seeds examples to train augmentation generation in the context of few-shot classification, and \newcite{yang2020generative} focused on low-resource in commonsense reasoning. 
Since the augmentation approach tried in those works is based on  straightforward training (fine-tuning)  a PLM-based text generator using the existing data (without exercising any topical control), we include the results from this general approach in ``aug baseline" column in addition to the baseline that does not attempt any augmentation (``off-topic" column in \autoref{tabResults}).
Since the above mentioned works also have demonstrated that classical ``back-translation" augmentation approach is substantially inferior to the PLM-based text generation, we decided not to include the former in our experiments.

\begin{figure}[!t]
\centering
\includegraphics[width=3.0in]{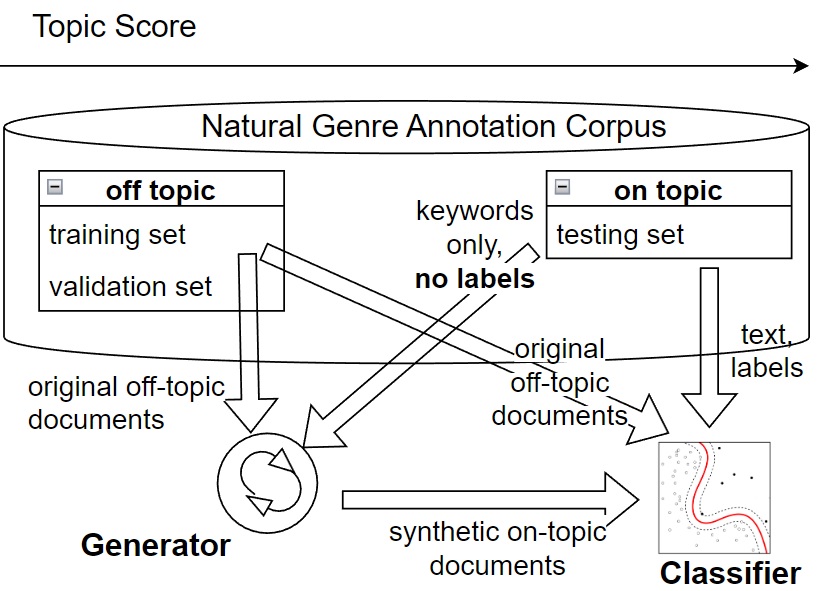}
\caption{Experimental workflow} 
\label{figDiagram}
\end{figure}

\newcite{jin2022deep} provides an overview of recent research in a closely related task of text style transfer (TST). Unlike TST, we are interested in keeping the topic, but not specifically concerned with preserving the {content} as long as the generated documents aid in domain transfer. 
The challenges maintaining coherent style and topic within longer texts (that exceed the current transformers' input limits of 500-4000 tokens) have been proposed to address by progressive generation \cite{tan20}. In this study, we are not as much concerned with quality of output texts, but rather with their help in domain adaptation. 

\section{Methodology}
\label{sec:Methodology}

Our study builds upon prior investigations into domain biases in text classification \cite{petrenz10,sharoff10lrec}, which largely depended on a limited set of hand-selected datasets with analogous tasks but varying distributions. We present a comprehensive methodology to assess and mitigate the domain transfer gap. The main idea is to simulate the situation when a classifier is trained on documents that lack a topic, e.g. \textit{medicine}, and then to test it on the documents where such topic is well represented. This performance is contrasted  with the situation when the classifier is initially trained on the documents where this topic is represented well. While our empirical results focus on genre classification, our methodology is directly applicable to other classification tasks such as gender, authorship, or sentiment classification. 

We train two classes of models: 
\begin{enumerate}[noitemsep]
    \item a topic model produced from a diverse corpus, even though it might be biased with respect to its genres, and 
    \item genre-classification models based on a PLM (such as Bert) which is fine-tuned on a genre-diverse corpus, even though each individual genre might be biased with respect to its topics. 
\end{enumerate}

Figure \ref{figDiagram} illustrates the overall workflow for our experiments.  

\subsection{Estimation of Topic Models} 
\label{sec:orgdf6063c}
{For our experiments, we needed as diverse topic model as possible so that we can assess the performance gaps when transferring between the topics}.
The topic model in this study was produced by a neural model \cite{dieng20topic} which can achieve better interpretability in comparison to traditional Latent Dirichlet Allocation (LDA) models \cite{blei03}.  More specifically, the Embedding Topic Model (ETM) differs from LDA by estimating the distribution of words over topics as:
$$w_{dn} \sim \text{softmax}(\rho^\top\alpha_{z_{dn}})$$
\noindent
where $\rho$ are word embeddings and $\alpha_{z_{dn}}$ are topic embeddings, {$dn$ refers to iteration over documents and topics, see} \citeauthor{dieng20topic} (\citeyear{dieng20topic}) for the full description of ETM.
For estimating the topic model, we used a topically-diverse corpus of ukWac \cite{baroni09} created by wide crawling of web pages from the .uk top level domain name (the total size of ukWac is 2 billion words, 2.3 million Web pages).  As suggested by \newcite{dieng20topic}, the number of topics of a topic model can be selected by maximising the product of topic coherence (the average pointwise mutual information of the top words for a topic) by its diversity (the rate of unique words in the top k words of all topics).  In this way we arrived at choosing 25 topics for the ukWac corpus, see \autoref{tabCorporaT}, Topic Coherence of this model is 0.195, Topic Diversity is 0.781.  In the absence of a gold test set for an unsupervised method, all of the topics are interpretable (the topic labels in \autoref{tabCorporaT}) have been assigned by inspecting the keywords and a sample of documents).
Topic 8 applies to short documents with residual fragments from HTML boilerplate cleaning in ukWac, so that the date and time indicators remain the only identifiable keywords for such documents.


\begin{table*}[t] 
\caption{Corpus of natural genre annotation \label{tabCorporaG}} 
\centering
\small
\label{GenreTable}
\setlength{\tabcolsep}{2pt}
\begin{tabular}{llrl|l}
Genre & General prototypes & Texts & Natural source & Bias (\autoref{tabCorporaT})\\
\hline
ARGument & Expressing opinions, editorials & 126755 & Hyperpartisan \cite{kiesel19semeval} & Topics 9, 13\\
INSTRuction & Tutorials, FAQs, manuals & 127472 & A sample of StackExchange & Topics 19, 21\\
NEWS & Reporting newswires & 16389 & Giga News \cite{cieri02} & Topics 5, 9\\
PERSonal & Diary entries, travel blogs & 16432 & ICWSM set \cite{gordon09} &  Topic 23\\
INFOrmation & Encyclopedic articles & 97575 & A sample of Wikipedia &  Topics 1, 15, 20\\
Review & Product reviews & 1302495 & Amazon reviews \cite{blitzer07} & Topics 1, 16, 17\\
\hline
 & \textbf{Total} & 1687118 & \\ 
\end{tabular}
\end{table*}

\subsection{Genre Corpus}
We also needed a corpus with good coverage of several genres. Up to our knowledge, there is no large corpus for that purpose, so we combined several data sources into a corpus of ``natural genre annotation'' so that each source is relatively homogeneous with respect to its genres.  The list of our genres follows other studies which detect text types which are common on the Web \cite{sharoff18genres}.  They
have been matched to commonly used datasets, such as a portion of the Giga News corpus to represent News reporting and the Hyperpartisan corpus to represent news articles expressing opinions.  The composition of the natural genre corpus is listed in Table \ref{tabCorporaG}. 
The corpus of natural genres is large, but it is biased with respect to its topics.  For example, the Amazon reviews dataset contains a large number of book and music reviews, and a small number of reviews of office products and musical instruments.  However, these are not the topics inferred by the topic model, as this division into topics exists only with the reviews dataset, while other sources of natural annotation have  no office products or musical instruments. What is more they are likely to have a very different structure of annotation labels even when there is some intersection between their topics.  For example, the category labels assigned to the pages in Wikipedia are different from both the Amazon review labels and for the inferred ukWac topics, while both  as listed in \autoref{tabCorporaT}. 
Having the topics for all sources as inferred by our topic model and the documents annotated with their genres gives two views on the same document,  for example, a document which starts with
\begin{examples}
   \small
 \item \label{egReviewE} There's little need to review this CD after Daniel Hamlow's thoughtful and informative critique above, but I loved the CD so much I had to weigh in.  In case you aren't familiar with his citations, he mentions the big three Brazilian music classics: Astrud Gilberto's "Jazz Masters 9" from Verve, "Jazz Samba" \ldots
\end{examples}
\noindent
can be described as a Review from its provenance from the Amazon reviews dataset and as primarily belonging to Topic 1 (Entertainment, \autoref{tabCorporaT}) from its ETM inference.

\begin{table*}[h]
\caption{Keywords from ukWac for the topic model with 25 topics \label{tabCorporaT}}
\centering
\small
\setlength{\tabcolsep}{2pt}
\begin{tabular}{lp{0.85\textwidth}}
Label: Nr & Top keywords \\
Finances: 0 & insurance, property, pay, credit, home, money, card, order, payment, make, tax, cost, time, service, loan\\
Entertain: 1 & music, film, band, show, album, theatre, festival, play, live, sound, radio, song, dance, songs, tv, series\\
Geography: 2 & road, london, centre, transport, park, area, street, station, car, north, east, city, west, south, council, local\\
Business: 3 & business, management, company, service, customers, development, companies, team, experience, industry\\
University: 4 & students, university, research, learning, skills, education, training, teaching, study, work, programme\\
Markets: 5 & year, market, million, energy, waste, years, cent, industry, investment, government, financial, increase\\
Web: 6 & information, site, web, website, page, online, search, email, click, internet, details, links, free, find, sites\\
Science: 7 & data, research, system, analysis, model, results, number, time, science, methods, surface, cell, energy, test\\
*Cleaning: 8 & 2006, 2005, posted, 2004, june, july, october, march, april, september, 2003, august, january, november, post\\
Politics1: 9 & government, world, people, international, war, party, countries, political, european, country, labour, british\\
Travel: 10 & hotel, room, day, area, house, accommodation, holiday, visit, city, centre, facilities, town, great, tour\\
Health: 11 & health, patients, treatment, care, medical, hospital, clinical, disease, cancer, patient, nhs, risk, drug\\
Councils: 12 & development, local, community, council, project, services, public, national, planning, work, government\\
Life1: 13 & people, time, questions, work, make, important, question, problem, change, good, problems, understand\\
Software: 14 & software, system, file, computer, data, user, windows, digital, set, files, server, users, pc, video, mobile\\
Sports: 15 & game, club, team, games, play, race, players, time, season, back, football, win, world, poker, sports, sport\\
Religion: 16 & god, life, church, people, lord, world, man, jesus, christian, time, love, day, great, death, faith, men, christ\\
Arts: 17 & book, art, history, published, work, collection, world, library, author, london, museum, review, gallery\\
Law: 18 & law, act, legal, court, information, case, made, public, order, safety, section, rights, regulations, authority\\
Nature: 19 & food, water, species, fish, plants, garden, plant, animals, animal, birds, small, dogs, dog, tree, red, wildlife\\
History: 20 & years, century, house, st, john, royal, family, early, war, time, built, church, building, william, great, history\\
Engineering: 21 & range, design, light, front, high, car, made, water, power, colour, quality, designed, price, equipment, top\\
Politics2: 22 & members, meeting, mr, committee, conference, year, group, event, scottish, council, member, association\\
Life2: 23 & time, back, good, people, day, things, make, bit, thing, big, lot, can, long, night, feel, thought, great, find\\
School: 24 & people, children, school, support, young, work, schools, child, community, education, parents, local, care\\
\end{tabular}
\end{table*}

\subsection{Transfer Assessment}
This subsection describes the methodology that we have developed to test the effect of a topic change. While this methodology is applicable to any non-topical classification, here, we describe how we use it with document genres. {Our main goal here is being able to create training, validation and test sets on particular topics to experiment with a genre classification task, specifically knowledge transfer between the topics}. We used the following procedure for estimating topical biases. For each topic as estimated by the topic model (e.g., ``Entertainment"), we create a dataset, that we label as \textit{off-topic}. For this, we take $N$ documents of each class (document \textit{genre} in our case).  For example, for N = 100  we take 100 argumentative texts, 100 instructions, 100 news reports, etc. such that the selected documents have the \underline{lowest} scores with respect to that topic, e.g. documents not about entertainment.

Through our experiments, we compare the classification results trained on  the \textit{off-topic} datasets with those trained on \textit{on-topic} datasets. The latter are constructed in exactly the same way except by selecting the documents with the  \underline{highest scores} on the topic, e.g. those \underline{most} relevant to entertainment. For each topic, we also created an \textit{on-topic} test set making sure it does not overlap with the training sets. Validation sets were off-topic since within a domain transfer setting there isn't any on-topic training data available. 
Specifically, in the experiments below, we used 300 documents of each genre in a test set, 300 documents of each genre in a validation set, and varied the sizes of the training sets as stated in our section \ref{sec:Results}.
This way we assess the ``domain transfer": a scenario when a model trained on \textit{off-topic} data needs to be applied to an \textit{on-topic} dataset.  
Structuring our datasets that way has several advantages: 1) both \textit{on-topic} and \textit{off-topic} sets have same number of documents in each class (genre) and the same total size, which allows us to determine the transfer gap under the same conditions, and 2) the datasets are automatically balanced with respect to each class (genre), even while our original corpus is not, thus the comparison metrics are more reliable and interpretable. 

To build the genre classifiers, we fine-tune the ROBERTA-large \cite{liu2019roberta} and BERT-large \cite{devlin2018bert} models from the Hugging-Face library\footnote{\url{https://huggingface.co/}} with the the common in the prior research learning rate of $10^{-5}$ for 6 epochs, using its Adam optimizer \cite{kingma2014adam}. 
Following the standard validation procedure, we report the F1 score computed on the respective test set for the number of epochs that showed the best score on the validation (development) set.


As a compromise between the reliability of our results and the processing time, after preliminary investigation we settled on working with the window of 1000 characters randomly positioned within a document.  Random positioning mitigates the impact of document structure, e.g. an introductory question positioned at the start of the StackExchange dataset. Our experiments with human raters show that the windows obtained this way still provide sufficient information to determine the topic and genre.

In order to mitigate the superficial differences between the sources, when training and applying our classifiers, we remove all the numbers and punctuation. We do not apply this filtering when training our text generators to preserve readability. We apply it to the generated texts instead.

\subsection{Data augmentation}
\subsubsection{Our Keyword Extraction Algorithm}
\hl{Our domain adaptation approach involves generating synthetic documents on a given topic. Thus, the generator is trained to receive a sequence of keywords and to generate a document in the desired category (\textit{genre} in this study).} We experimented with several variations of a heuristic algorithm to select the keywords and settled on the following approach after manually inspecting the quality of the generations and their topical relatedness.  We are not much concerned how truthfully the keywords represent the content of the document, but rather how well they represent the \underline{topic} to enable topic-focused generation. Thus, when deciding which words to extract as keywords, we promote those that are strong representatives of the document topic, which is quantitatively assessed by our topic model. It assigns each word (in the corpus) a score with respect to each topic between 0 and 1. The higher the score the stronger the word is related to the topic. Since some documents mix several topics, at  times with numerically similar proportions, we accordingly weight the individual word scores with the overall topic scores in the document. Finally, we also want to adjust for repeated occurrences of the same word. Thus, our word scoring formula (within a document) simply iterates through all the topics and through all the word occurrences in the document while adding up the word scores with respect to the corresponding topic:
$$score(w,D) = \sum _{i \in {D _w}} \sum _{t} L(D,t) \cdot L(w, t)$$

\noindent
where $i$ goes over all the occurrences of the word $w$ in the document $D$, $t$ goes over all topics (25 in the study here), $L(D,t)$ is the score of the document with respect to topic $t$ and $L(w, t)$ is the score of the word $w$ with respect to topic ${t}$.

We preserve only 10 top-scoring words in each document, so all the other words are discarded and the original sequence of the remaining words becomes the keyword sequence for the generator. Table \ref{GenExamples} in Appendix shows an example of extracted keywords along with how they are used to generate new synthetic documents, as detailed in the following subsection.




\subsubsection{{Our Topical Augmentation Control}} 
Our suggested method of improving domain transfer proceeds by augmenting the \textit{off-topic} training set with automatically generated \textit{on-topic} documents. Thus, in a practical scenario, the test topics (keywords) don't have to be known in advance but can be extracted from previously unseen test documents from the target domain. The only tool required for this is an existing topic model, which can be built similarly to as we did here on any general corpus of a modest size, e.g. two billion words of ukWac, \cite{baroni09}, which is not resource-consuming.

To achieve this we fine-tune a pre-trained language model into a separate generator for each of our genres (listed in \autoref{GenreTable}).  Our earlier experimenting with using a single model for all genres and a special token to specify the desired genre resulted in weaker results. 
For this fine-tuning,  we use exactly the same $N\cdot6$ documents as are in our \textit{off-topic} training set, thus operating in a practical scenario when on-topic documents are not available.  Each generator is fine-tuned to take a sequence of keywords extracted according to the algorithm detailed above as input and to generate a document in the genre corresponding to this generator and of the topic defined by the keywords. During fine-tuning, the generators learn to associate the input keywords with the content of the output document, { which becomes an important mechanism of topic control and facilitating the domain transfer.}

We specifically used T5 as our generating model \cite{raffel20}. 
It is a unified text-to-text transformer, trained on the Colossal Common Crawl Corpus to predict the next word based on the preceding words in an auto-regressive way.  We used the small version since we did not observe any advantage in using the Base or Large T5 model in our early experiments, so we kept the less computationally intensive model.  Its input format requires a prefix to indicate which downstream task is being fine-tuned, so we used the word ``generate.'' We trained each model for 16 epochs using Simple Transformers library\footnote{https://simpletransformers.ai/} with a default learning rate of $.001$ and its Adam optimizer. 
For generating, we also use the following T5 hyper-parameters, specifically the number of beams $=1$, top $k=50$, top $p=.95$. The selected hyper-parameters were chosen after preliminary experimentation by inspecting the produced quality of generations in terms of both topical and genre fit. Table \ref{GenExamples} in the Appendix illustrates our domain adaptation approach by examples of extracted keywords and synthetic documents generated from those keywords in different genres. 

One of our overall hyper-parameters is how many documents to generate. Our preliminary experimentation suggested that 1:1 was a near optimal ratio: the same number of original and synthetic documents. We include several other combinations in our empirical results below. 
\begin{table*}[htbp]
\caption{{Averaged across topics F1 scores for testing genre classification domain transfer gaps and our augmentation approach. The ``on-topic" columns show the performance when training and testing on in-domain documents. The ``off-topic" columns present training on the off-topic documents and testing ``on-topic". ``aug baseline" is the result of augmentation by generation without domain adaptation.
Our domain adaptation augmentation results are in the last column for each N (``aug adapt"). The results for separate topics are included in the Appendix. All our results are statistically significantly different from the baselines at the level $p<0.01$. There is no statistically significant difference at that level between ``aug baseline" and ``off-topic".}
\label{tabResults}}
\setlength{\tabcolsep}{4pt}
\centering
\small
\begin{tabular}{l|llll|llll|llll}
\label{MainResults}
 &  & N=30 &  & & & N=100 &  &  & N=1000 & \\
Topics: & on-   & off-  & aug      & aug & on- & off- & aug & aug &  on- &off  & aug & aug\\
        & topic & topic & baseline & adapt         & topic & topic & baseline & adapt &  topic & topic  & baseline & adapt        \\

\textbf{Model}:  & &  &  &  & & & & & & & & \\ 
\textbf{Roberta Large}  & 77.0 & 52.1 & 52.4 & \textbf{58.1} &  & 61.2 & 61.4 & \textbf{65.1} & 94.9 & 78.8 & 79.0 &  \textbf{81.7} \\ 
\textbf{Bert large}& 79.5 & 49.9 & 51.1 & \textbf{55.2 } &  84.9 & 55.6& 55.0 & \textbf{58.8} & 85.1 & 68.8  & 68.7 & \textbf{71.9} \\

\end{tabular}
\end{table*}

\section{Experiments}
\label{sec:Results}
The most time-consuming part of our experiments were fine-tuning the generators (T5)  and the classifiers at the cost of roughly 6000 hours of NVIDIA GeForce RTX 2080.
\subsection{Comparison Results}
We assess the effect of domain mismatch and our approach to improving domain {transfer} by augmenting the training sets with synthetic on-topic documents. The difference between the accuracy obtained before and after generation demonstrates the efficiency of the augmentation model. Table \ref{MainResults} shows the comparison results for 3 different sizes of training data: 1000, 100 or 30 documents per genre. As we can see, the topic mismatch effect is \textit{extremely significant}: the average absolute F1 drop from \textit{on-topic} to \textit{off-topic} training set is around 20\% for N = 1000 and 30\% for smaller Ns. The average \textit{on-topic} F1 score  for the largest size is 86.4\%, while in our tests the human raters achieved 93\% on a sample of 100 documents of each genre. 
The average \textit{off-topic} performance for that size drops to 66.8\%.
{All three configurations (``aug adapt" columns) demonstrate 2-6 percentage point increases in F1 from non-augmented off-topic training sets (``off-topic" columns). At the same time, the straightforward ``augmentation by generating" approach from prior works (``aug baseline" columns) does not show any noticeable improvement, even though it was found by prior work somewhat effective in several tasks not involving domain transfer. 
We hypothesise that this is because the general approach does not provide a mechanism to facilitate \textit{domain transfer}, while our approach does. 
All the differences between our approach and the baselines are statistically significant at the level of alpha $0.01$ according to a pairwise t-test. This confirms empirically with high confidence that our augmentation procedure is beneficial for genre classification.} 
\hl{While in this current study we prioritized reporting metrics averaged across all 25 topics rather than on individual topic level, we still can observe that the magnitude of the transfer gap and the augmentation effects are normally consistent across all the configurations and models used, see }\autoref{tabResultsFull} in the Appendix. Still, there are some exceptions due to a large number random factors involved including the choice of off-topic documents, the quality of synthetic documents in terms of both genre and topic, the optimality of hyperparameters, and others.

Qualitative analysis demonstrates that little recovery is possible in case of a very strong correlation between the topics and genres, for example, scientific texts (Topic 7) mostly occur in the genre category of Academic texts; similarly, texts related to law (Topic 18) mostly occur in News reporting.  The quality of generation in these topics for other genres remains low.

\subsection{Ablation Studies}
This subsection reports several \textit{ablation} experiments that we conducted to additionally verify the effects reported above and to gain the insight into the phenomena studied. In order to verify that the genre labels in our synthetic texts were important we randomly shuffled their labels. This way, the augmented data became to act only as noise. Not surprisingly,  the average scores dropped to the baseline levels which verified that using the proper model for each genre to generate the synthetic augmenting texts is important, and that the improvements reported above were not due to simply the change in the statistical properties of the training and validation sets or due to addition of noise.

We also looked at several ways of mixing the original and augmented data. Table \ref{Ablations} 
presents the average  across topics scores for various sizes used. 
It can be observed that while some small improvements can be achieved by generating more documents, those gains are not statistically significant. On the other side, very small numbers of added documents indeed result in statistically detectable drops. Using only synthetic documents results in drops to the levels only slightly above or even below baselines. \hl{We also observed that using keywords from randomly selected off-topic documents is significantly worse than using those from the on-topic documents, which confirms that using domain adaptation mechanism such as suggested here by us is crucial. The details are in the last rows for each N in Table }\ref{Ablations} in the Appendix. 

{We have also looked at the optimal choice of the number of keywords. While the details are presented on Figure }\ref{fig:plot} \hl{in the Appendix, it is worth noting here that the optimal number is indeed around 10-20 keywords. Also, the augmentation affect drops to 0 on both ends: Too few keywords means no topical control is performed. 100+ keywords result in practically all the non-stop words treated as keywords. This means the model does not really learn how to generate a document on a topic specified by a set of keywords but it rather learns how to restore deleted stop-words from the given text.} 

\subsection{Qualitative Exploratory Study with ChatGPT}
As a further qualitative investigation into the problem, we have also confirmed that a much larger language model still suffers a domain transfer gap when tasked with genre classification. We have randomly sampled 72 triples consisting of a pair of non-identical genres and a topic. Then, we compared binary classification accuracy by entering specially crafted prompts into ChatGPT\footnote{Accessed throughout March-April 2023}, which is built on top of GPT-3.5 model with approximately 355 billion parameters. 
An example of prompts is presented in \autoref{GPTExamples} in the Appendix. Each includes 5 randomly selected document examples of each genre (5-shot). 
The choice of those numbers was dictated by the combination of input size limitation, our early experience and prior studies on text classification with ChatGPT. 
For assessing a domain transfer gap, we followed the same methodology as described in section \ref{sec:Methodology}: we compared the binary classification performance when \textit{off-topic} documents were used as prompt examples with when \textit{on-topic documents} were used. We have indeed verified that the domain gap exists even in a language model of that size: the average accuracy with on-topic examples was 83\% while the average accuracy when using off-topic examples was 42\%. We also estimated human accuracy in this setup as 88\%.

When experimenting with our prompts, we discovered that it was crucial to use chain-of-thought (CoT) approach (e.g.\citeauthor{wei2022chain}, \citeyear{wei2022chain}): 
After presenting examples of both classes, we  asked the model to ``list at least three criteria by which Class 1 and Class 2 texts are different from each other." Examples of the criteria generated by the model can be found in Table \ref{GPTCriteiaExamples} in the Appendix. We have qualitatively (informally) observed that:
1) ChatGPT was able to use both on-topic and off-topic examples to produce criteria that looked potentially useful for genre classification, e.g. ``Class 1 texts appear to be informational or factual, whereas Class 2 texts appear to be more conversational or personal in nature." or ``Class 1 texts are typically more objective and neutral in tone, while Class 2 texts tend to be more subjective and expressive." 2) Both on-topic and off-topic examples occasionally resulted in the criteria that are topic-reliant, e.g. ``Class 1 texts provided are about musicians and their careers" or "Class 2 uses words like \textit{position, certified gold,} and \textit{innovation}." 3) The presence of topically-reliant criteria was stronger with off-topic examples.

Next, within our prompt, we separately asked to apply each of the three criteria to the given test document, followed by a request to combine the criteria to make a classification decision. Examples can be found in Table \ref{GPTCriteiaApplicationExample} in the Appendix. By inspecting the model's responses, we have observed that using off-topic examples resulted in the following types of chain-of-thought ``confusion" to happen more often than using on-topic examples: 1) applying not the same criteria that originally stated 2) applying a criterion incorrectly. 3) erroneously ``swapping" classes when combining. 
This suggests that while ChatGPT has strong ``emerging" capabilities for recognizing genres (see another confirmation at \newcite{kuzman2023chatgpt}), they are weaker when the examples are off-topic and so are more likely to ``break" the chains of thoughts.

\section{{Conclusions}}  
\label{sec:org7a07b14}
We have demonstrated a severe degradation in a PLM-based \textit{document} classifier when trained on one topic, such as \textit{politics}, and tested on another, like \textit{healthcare}. Rather than following the prior empirical studies on the impact of domain transfer  that involved only a few hand-picked datasets with similar tasks but somewhat different data distributions, we have developed a methodology based on a neural topic model to assess the domain transfer gap between a wide variety of topics. While our empirical results focus on genre classification, our methodology is applicable to other classification tasks such as gender, authorship, or sentiment classification. We have also shown that the topic transfer gap can be mitigated by means of proper topic control while generating additional training documents (augmentation). As a result of our approach, a model to predict a non-topical category (genres in the case here) can be trained on the documents in one topic (e.g. \textit{politics}) and applied to another (e.g. \textit{healthcare}) even when there are no healthcare-related documents in the training corpus.} 
{We have also created a large corpus with natural genre annotation and a very general/diverse topic model. Both can be used in follow-up studies.} 

{Still, our study has certain limitations. The degree of improvements from augmentation is not uniform. For some topics we obtain much better results than for others, while occasionally the performance on the augmented set is even lower than on the original off-topic training set. This is likely to be related to the high degree of correlation between the topics and genres, for example, the lack of texts on the topic of law in genres other than news reporting in our corpus, thus leading to less successful attempts to generate discussions, academic articles or advice texts on this topic. We need to find better ways to improve off-topic generation when {it makes no positive impact on the accuracy of classification of on-topic test texts}, possibly by using very large language models.

Nevertheless, through a qualitative exploratory study with ChatGPT we were able to confirm that even such larger language models still suffer from the domain transfer gap. Even while our approach does not solve this very challenging domain transfer problem completely, it suggests a direction in which a small but productive step can be made. Larger pre-trained language models can be tried in future such as GPT-4, for both  generation and classification.} 

{Also, larger training sets can be explored, as well as in ``few-shot" settings. A number of approaches improving the quality of generated text, e.g. those based on Generative Adversarial Networks} \cite{goodfellow2020generative} or meta learning \cite{lee22meta} can be explored, {as well as various methods to control the quality and topical fit of the generated texts.}

\section*{Limitations}

We have already discussed several limitations of our study in the preceding section. Since our primary focus was on reporting metrics averaged across all 25 topics, this approach prevented us from discerning clear patterns or relationships between the properties of individual topics, domain gaps, and the effects of augmentation. More research is needed to investigate topic-level conditions for successful transfer. We intend to address this in future work. Given the computationally demanding nature of our experiments, we have limited our study to short text samples rather than full documents. Our corpora consisted exclusively of English documents, which might limit empirical findings to languages with limited morphological complexity. While we utilized Latent Dirichlet Allocation, other topical models might also be suitable for assessing domain transfer, and alternative augmentation methods might be worth exploring. For better generalization, a corpus with a larger set of genres can be assembled and explored. Additionally, other tasks such as authorship or sentiment classification could be explored in this context.

\bibliographystyle{acl_natbib}  
\bibliography{bibexport}

\appendix 
\section{{Appendix}}

\begin{table*}[h]
\caption{{Domain Adaptation}: examples of documents generated in different genres from the same keywords on the topic of Entertainment (topic 0 in Table \ref{tabCorporaT}). Keywords (shown in \textit{italic}) are taken from an ``on-topic" document in a test set, but its genre label (product review in this case) is not used since the task is to predict it. Instead, synthetic documents are generated from those keywords in each genre. Those documents augment the training set. The generated documents respect the constraints of topics, keywords and genres.} 
\small
\begin{tabular}{|p{.99\textwidth}|}
\hline
\label{GenExamples}
\textbf{On-topic Review document:} As a serious \textit{form} of \textit{music} outside of Jamaica, on par with American and British rock and roll and R\&B, Before this breakthrough   Reggae was seen as something of a joke  even though Jimmy Cliff’s  ``The Harder They Come" \textit{set} the stage ``Catch a \textit{Fire}"    clinched it.    The \textit{album} is a solid classic and a masterpiece with lead vocals not just from Marley but band-mate Peter Tosh as well    on the tracks     \textit{years}    and    stop that train       the deluxe edition features both the \textit{original}   ``Catch a \textit{Fire}"    \textit{album}    released only in Jamaica in December of with its two bonus tracks as well as the \textit{original} \textit{album} released internationally to the \textit{public}. \\
\textbf{keywords:} \textit{form music set fire album years original fire album original album public date album album fire} \\
\textbf{Generated texts} for the topic of Entertainment with these keywords:\\
\textbf{Information:} He wrote poetry in a Broadway, and was a semi - finalist. As a performer, Eve was renowned for his strong wit and strong vocals, as well as his striking delivery of heavy metal, a set of wits and his physical presence on stage, earned him the title ``Master Of Song". Eve was also the master of wits' composition as well as some of his later works. He opened \textit{fire} to wits in. Eve appeared in two of his first two albums as a performer and then mixed wits and wits, gaining his first \textit{album}, a release from his old band. Eve was the sound of a song for eve, which remained unfinished. \\
\textbf{Instructive discussion:} Do you have a video on repeat since the original hits them on the radio every evening for the first two pieces sold separately by the label? Does anyone have an official release \textit{date} or should i be? Thank you in advance. 
On this \textit{date} on your public air \textit{date} the entire \textit{album} is sold separately, which is released through all tv series. the original release \textit{date} is released on the public register for all tv series and can be purchased on every \textit{album}. \\
\textbf{Product review:} He was the king of rock and roll, legendary guitarist Peter Tosh, and drummer Neal Schon. Of course, if you really like your \textit{music}, then you should buy this \textit{album}. from this CD, you'll see why. from the infectious, surrealistic guitar riffs that set the stage, at the beach or in the hall of the ancient mariner, to the funky ``Catch a \textit{Fire}", to the insanely catchy catchy summer pop staple. \\ 
\textbf{Argumentation:} I'm an atheist myself, and I'm just a British freak! maybe if it was only about pop culture then it would be rude to you to leave the UK, and do something about it. I don't feel like am American player doing the same with a \textit{Fire-}fox. You can add to that one if you're serious about it. I don't hold an open exchange about anything. I was an apologist to the \textit{album} you were talking about. \\
\hline
\end{tabular}
\end{table*}

\begin{table*}[h]
\caption{F1 score results for fine-tuned Roberta Large classifier reported by separate topics. The ``on-topic" columns show the performance when training and testing on in-domain documents. The ``off-topic" columns present training on the off-topic documents and testing ``on-topic". Our augmentation results are in the last column for each N.
\label{tabResultsFull}}
\setlength{\tabcolsep}{4pt}
\centering
\small
\begin{tabular}{l|rrr|rrr|rrr}
\label{MainResultsDetails}
 &  & N=30 &  &  & N=100 &  &  & N=1000 & \\
Topics: & on-topic & off-topic & aug adapt & on-topic & off-topic & aug adapt &  on-topic &off-topic  & aug adapt\\
Finances: 0 &73.9 & 52.0 & 62.0 &94.2 & 60.2 & 65.2 & 95.6 & 84.3 & 83.1 \\
Entertain: 1 &76.8 & 51.1& 61.2& 96.0 & 65.0 & 63.0 & 97.1 & 79.0 & 79.5 \\
Geography: 2 &84.2 & 55.5& 73.9& 95.5 & 65.1 & 64.6 & 97.3& 95.3 & 93.1 \\
Business: 3 &81.9 & 49.1& 52.3& 93.8 & 56.7 & 67.6 & 97.2& 93.8 & 88.5 \\
University: 4 &80.5 & 58.5& 61.3& 89.2 & 67.7 & 71.2 & 97.5& 81.0 & 83.2 \\
Markets: 5 &78.9 & 46.6& 47.3& 87.9 & 50.0 & 53.7 & 89.0& 66.8 & 70.8 \\
Web: 6 & 74.6& 42.7& 56.5& 92.2 & 57.4 & 64.5 & 97.2 & 94.5 & 93.1 \\
Science: 7 & 78.1& 56.6& 54.7& 86.6 & 61.5 & 66.0 & 92.3 & 82.3 & 84.0 \\
*Cleaning: 8 & 76.8& 63.3& 73.0& 84.5 & 66.2 & 72.2 & 93.0 & 73.7 & 73.2 \\
Politics1: 9 & 66.7& 39.7& 58.3& 85.1 & 51.0 & 56.9 & 92.4 & 59.4 & 63.5 \\
Travel: 10 & 89.5& 65.5& 71.0& 93.1 & 63.4 & 71.7 & 96.6 & 69.1 & 81.8 \\
Health: 11 & 72.7& 50.1& 57.3& 88.0 & 57.2 & 61.1 & 85.0 & 69.2 & 67.3 \\ 
Councils: 12 & 82.0& 49.6& 53.4& 94.2 & 58.9 & 67.0 & 95.5 & 87.2 & 87.5 \\
Life1: 13 & 80.6& 45.6& 54.1& 94.5 & 45.4 & 57.2 & 97.0 & 89.9 & 84.8 \\
Software: 14 & 85.8& 52.4& 52.8& 94.7 & 64.1 & 65.0 & 96.0 & 62.3 & 79.1 \\
Sports: 15 & 88.4& 49.0& 75.5& 93.4 & 70.2 & 64.6 & 94.3 & 65.9 & 79.3 \\
Religion: 16 & 73.2& 48.9& 66.6& 87.0 & 56.0 & 64.8 & 95.7 & 79.4 & 80.4 \\ 
Arts: 17 & 72.1& 39.8& 65.1& 89.6 & 72.0 & 76.3 & 94.5& 66.6 & 69.1 \\ 
Law: 18 & 65.5& 51.3 &  47.0 & 81.2 & 51.0 & 56.0 & 93.0& 75.2 & 83.2 \\
Nature: 19 & 77.1& 55.9& 61.1& 94.1 & 65.0 & 75.1 & 95.3& 81.4 & 93.4 \\
History: 20 & 69.2& 49.6&51.3  & 88.7 & 64.7 & 62.2 & 93.9& 86.6 & 84.3 \\
Engineering: 21 & 86.6& 55.0& 47.5 & 96.8 & 66.3 & 64.7 & 97.7& 86.2 & 93.0\\
Politics2: 22 & 71.2& 49.6& 55.7& 85.8 & 62.2 & 63.1 & 94.9& 66.4 & 67.0 \\
Life2: 23 & 75.4& 47.1& 47.2& 93.5 & 66.0 & 66.6 & 96.5& 73.4 & 83.3 \\
School: 24 & 73.2& 59.1& 60.2& 94.8 & 64.6 & 78.3 & 97.2& 83.8 & 85.5 \\

\hline  
Average & 77.0 & 52.1 & 58.1 & 91.0 & 61.2 & 65.1 & 94.9 & 78.8 & 81.7 \\
\end{tabular}
\end{table*}

\begin{table*}[htbp]
\caption{Ablations: average performance for mixing original and synthetic documents. The statistical differences at the level of .05 from the best configuration within each N are marked with \textsuperscript{++}.} 
\centering
\small
\begin{tabular}{llll}
\label{Ablations}
\textbf{Original} & \textbf{Augmented} &  \textbf{Roberta Large} & \textbf{Bert Large} \\ 
\hline 
\textbf{1000}  & 0 (baseline) & 78.8\textsuperscript{++} & 68.8\textsuperscript{++} \\  
\hline
{1000} & 10 & 79.0\textsuperscript{++} & 69.1\textsuperscript{++} \\  
{1000} & 100 & 80.1\textsuperscript{++} & 70.8\textsuperscript{++} \\  
{1000} & 1000 & 81.7  & 71.9 \\  
{1000} & 3000 & \textbf{81.9} & \textbf{72.0} \\  
{1000} & 5000 & 81.5 & 71.8 \\  
{0} & 1000 & 78.6\textsuperscript{++} & 68.9\textsuperscript{++} \\  
{1000} & 1000 random & 78.6\textsuperscript{++} & 68.5\textsuperscript{++} \\  
\hline 
\textbf{100}  & 0 (baseline) & 61.2\textsuperscript{++} & 55.0\textsuperscript{++} \\  
\hline 
100 & 10 & 61.3\textsuperscript{++} & 56.3\textsuperscript{++} \\  
100 & 100 & 65.1 & \textbf{58.8} \\  
100 & 300 & \textbf{65.4} & 58.7 \\
100 & 500 & 65.2 & 58.7 \\  
0 & 100 & 60.8\textsuperscript{++} & 54.7\textsuperscript{++} \\  
100 & 100 random & 61.4\textsuperscript{++}  & 55.3\textsuperscript{++} \\  
\hline
\textbf{30} & 0 (baseline) & 52.1\textsuperscript{++} &  49.9\textsuperscript{++} \\  
\hline
30 & 10 & 56.9\textsuperscript{++} & 53.3\textsuperscript{++} \\  
30 & 30 & 58.1 &  \textbf{55.2} \\  
30 & 100 & \textbf{58.3} & 55.1 \\
30 & 150 & 58.2 & 55.1 \\
0 & 30 &  52.9\textsuperscript{++} &  51.2\textsuperscript{++} \\  
30 & 30 random & 52.4\textsuperscript{++}  & 50.8\textsuperscript{++} \\  
\end{tabular}
\end{table*}

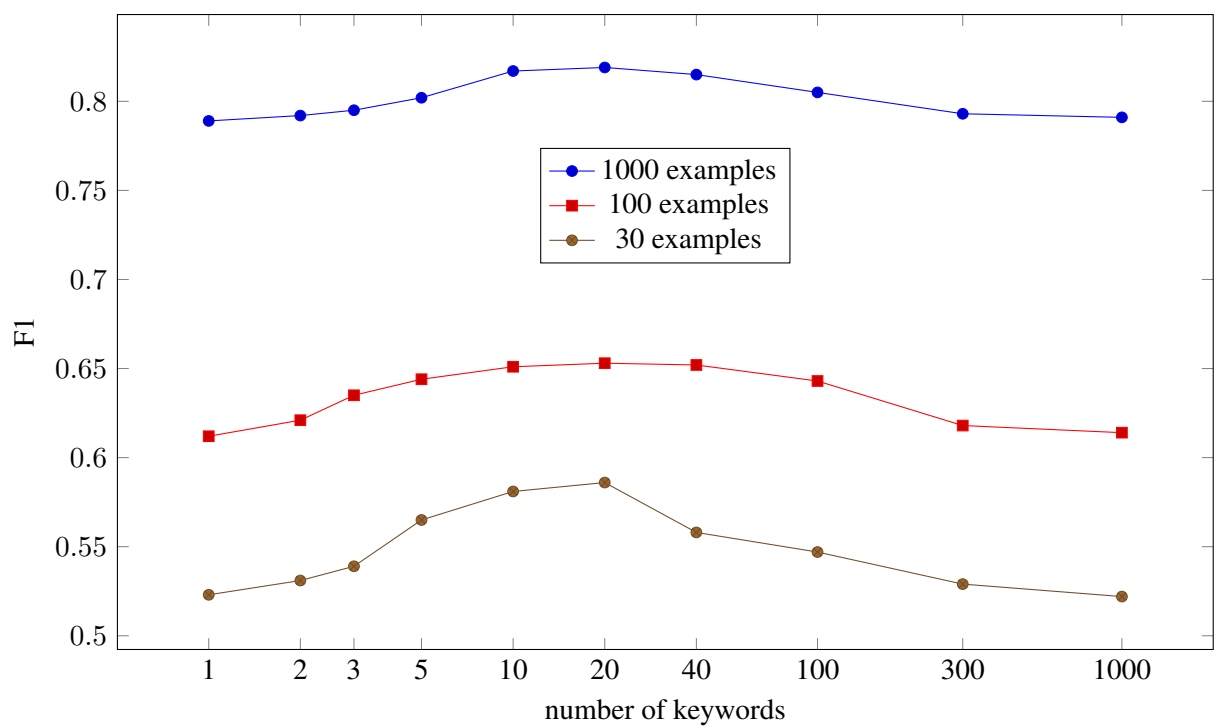
\begin{figure*}[htb]
  \centering
  \begin{tikzpicture}
  \begin{axis}
[
      xlabel     = number of keywords, 
      ylabel     = F1,
     clip       = false, 
     xmode=log, 
     height=10cm,
     width=\linewidth,
    legend style={at={(0.5,0.79)},anchor=north},
    xtick={1,2,3,5,10,20,40,100,300,1000}, 
    xticklabels={1,2,3,5,10,20,40,100,300,1000}    
]
    \addplot+[sharp plot] coordinates 
      { (1,.789)(2,.792)(3,.795 )(5,.802 )(10,.817 )(20,.819 )(40,.815)(100,.805) (300,.793)(1000,.791)}; 
    \addplot+[sharp plot] coordinates 
      { (1,.612)(2,.621)(3,.635)(5,.644)(10,.651)(20,.653)(40,.652)(100,.643) (300,.618)(1000,.614)}; 
    \addplot+[sharp plot] coordinates 
      { (1,.523) (2,.531) (3,.539 ) (5,.565 ) (10,.581 ) (20,.586 ) (40,.558) (100,.547)(300,.529) (1000,.522)}; 
\addlegendentry{1000 examples}
\addlegendentry{100 examples}
\addlegendentry{30 examples}
  \end{axis}  
  \end{tikzpicture}
  \caption{{F1 metric of performance for various numbers of keywords and data sizes with Roberta Large classifier.}}
  \label{fig:plot}
\end{figure*}

\begin{table*}[h]
\caption{Example of ChatGPT prompts used in our study. Class 1 is Information. Class 2 is News reporting. The topic is "Entertainment". Off-topic class examples.}
\small
\label{GPTExamples}

Based on the examples of texts of Class 1 and texts of Class 2 below, list at least three criteria by which Class 1 and Class 2 texts are different from each other. 

Here are some example texts of Class 1:

Example 1: World Darts Championship: He defeated number five seed Tony Eccles in the first round but lost to Shaun Greatbatch in round two. PDC career: Laursen became the first Dane to play in the PDC World Darts Championship. In the competition, he beat Colin Monk in the first round but lost to Dennis Priestley in the second round. Despite the fact that Laursen was up and missed eight darts to win the match before losing. He came through the Danish qualifying system for the second time for the PDC World Darts Championship but lost to Alan Tabern in the first round. Laursen has had some success in tournaments in his own country, reaching the final of the Danish Open (losing to Vincent van der Voort) and winning the Danish National Championships in 20. Laursen once again represented his country in the PDC World Darts Championship, having

...

Here are some example texts of Class 2: Liverpool, Manchester United, Arsenal, and West Ham in recent weeks, at least finished the half on a high. Blackburn captain Tim Sherwood just shot past the left-hand post in the 33rd minute after breaking through from a deep position and receiving an accurate pass from Jason Wilcox. After Asprilla shot over the bar and saw another effort pushed away by Flowers, Blackburn had another superb opportunity from Sherwood in the 38th minute. Wilcox again fed Sherwood, but his powerful shot could only find the crossbar via a deflection. Then Batty received a square pass from the right from substitute Keith Gillespie before firing home with a rare left-foot shot into the right-hand corner of Flowers' goal. Then, four minutes from time, Shearer fed Graham Fenton who charged into the area and volleyed first time past Hislop, who could only knock the ball high into the net. With a draw seemingly on the cards in the dying seconds, Fenton broke through once more to chip Flowers

...

Now, apply the first criteria to the following text:

Kuhn originated the role of Helen Bechdel in the off-Broadway Public Theater production of the musical Fun Home, which began its run September and opened officially on October. The run was extended multiple times and closed on January. She played the same role in the Broadway production, which ran from April to September at the Circle in the Square. Kuhn played the role of Golde in the Broadway revival of Fiddler on the Roof, starting on November. She plays Golde in the Menier Chocolate Factory London production of Fiddler on the Roof, which began on December and runs to March. Her television credits include Law \& Order and Law \& Order: SVU, All My Children, and two PBS shows: My Favorite Broadway: The Leading Ladies, recorded, released and in performance at the White House, and A Tribute to Broadway. The shows in March, Kuhn sang

Now, apply your second criteria to the same text.

Now, apply your third criteria to the same text.

Now, combine the criteria to decide which of those two classes the same text is more likely to belong.

\end{table*}

\begin{table*}[h]
\caption{Examples of criteria generated by ChatGPT. Class 1 is Information. Class 2 is Personal blogs. }
\small
\label{GPTCriteiaExamples}
Class 1 texts appear to be informational or factual, whereas Class 2 texts appear to be more conversational or personal in nature. Based on the examples provided, here are three criteria that differentiate Class 1 and Class 2 texts:

Language style: Class 1 texts tend to use more formal language and technical jargon, while Class 2 texts tend to use more informal language, colloquialisms, and slang. For example, Class 1 uses words like "position," "certified gold," and "innovation," whereas Class 2 uses phrases like "thrown down a little twist," "gone crazy," and "sooth your soul."

Content: Class 1 texts are often more focused on specific topics or subjects, while Class 2 texts tend to be more varied in content and topic. For example, the Class 1 texts provided are about musicians and their careers, while the Class 2 texts cover a range of topics, such as the Olympics, Chinese music, and personal relationships.

Tone: Class 1 texts are typically more objective and neutral in tone, while Class 2 texts tend to be more subjective and expressive. For example, the Class 1 texts provide information about musicians and their achievements without expressing personal opinions, while the Class 2 texts express personal feelings and opinions, such as excitement about the Olympics or admiration for Chinese music.

\end{table*}

\begin{table*}[h]
\caption{Examples of ChatGPT applying a criteria generated previously to a test document from the category of INFOrmation. Class 2 corresponds to Personal blogs. } 
\small
\label{GPTCriteiaApplicationExample}
\textbf{Criteria Generated}:

Language style: Class 1 texts tend to use more formal language and technical jargon, while Class 2 texts tend to use more informal language, colloquialisms, and slang. For example, Class 1 uses words like "position," "certified gold," and "innovation," whereas Class 2 uses phrases like "thrown down a little twist," "gone crazy," and "sooth your soul."

\textbf{Test text}:

Orchestras of Ray Ventura and Jacques Hélian, Adison's band, often billed as Fred Adison and his Collegians, was one of the principal French backing groups for singers and films in the 1930s. Adison also recorded copiously and released many RPM commercial recordings during this time. After the onset of World War II, he toured with Django Reinhardt in September-October 1939 and continued writing music for film. He was imprisoned in a Nazi war camp in 1940. Following the war, he led a new big band. In 1956, Charles Spiessert, owner of Cirque Pinder, hired Fred Adison to conduct the circus orchestra, a position he held until 1964. He accompanied Luis Mariano and Gloria Lasso during this time and recorded several albums of circus music. Following this, Adison went into semi-retirement, occasionally playing with small ensembles and dance orchestras on television. Filmography: Chantal Brunschwig, Louis-Jean Calvet, Jean-Claude Klein, "Cent Ans De Chanson."

\textbf{Generated output when the criteria applied}:

Based on the application of Criteria 1, which is language style, the text appears to belong to Class 1 - informational or factual, as it is written in a formal style and uses technical terms and proper nouns related to music and history.

\end{table*}

\end{document}